\def\year{2021}\relax
\documentclass[letterpaper]{article} 
\usepackage{utils/aaai21}  
\usepackage{times}  
\usepackage{helvet} 
\usepackage{courier}  
\usepackage[hyphens]{url}  
\usepackage{graphicx} 
\usepackage{graphicx}  
\usepackage{natbib}  
\usepackage{caption} 

\usepackage{amssymb} 

\usepackage{enumitem}
\usepackage{booktabs}

\usepackage{amsmath}
\usepackage{xspace}

\usepackage[dvipsnames,table]{xcolor}
\usepackage[pagebackref=true,breaklinks=true,colorlinks,bookmarks=false, citecolor=ForestGreen]{hyperref}
\usepackage{color, colortbl}
\usepackage{cleveref}

\usepackage[switch]{lineno}

\crefformat{section}{\S#2#1#3} \crefformat{subsection}{\S#2#1#3}
\crefformat{subsubsection}{\S#2#1#3}
\crefformat{table}{Table~#2#1#3}
\crefformat{tables}{Tables~#2#1#3}
\crefformat{figure}{Figure~#2#1#3}

\definecolor{GREEN}{RGB}{0,255,0}

\definecolor{anchor}{rgb}{0.5, 0.0, 0.5}
\definecolor{positive}{rgb}{0.5, 0.5, 0.0}
\definecolor{negative}{rgb}{0.0, 0.3, 0.8}

\title{Enhancing Unsupervised Video Representation Learning \\ by Decoupling the Scene and the Motion}
\author {
        \small
        Jinpeng Wang\textsuperscript{\rm 1, \rm2}\thanks{The first two authors contributed equally. This work was done during Jinpeng Wang's internship at Tencent Youtu Lab.},
        Yuting Gao\textsuperscript{\rm 2}\footnotemark[1],
        Ke Li\textsuperscript{\rm 2},
        Jianguo Hu\textsuperscript{\rm 1},
        Xinyang Jiang\textsuperscript{\rm 2},
        Xiaowei Guo\textsuperscript{\rm 2},
        Rongrong Ji\textsuperscript{\rm 3}, Xing Sun\textsuperscript{\rm 2}\thanks{Corresponding Author} \\
}
\affiliations {
    \textsuperscript{\rm 1} Sun Yat-sen University
    \textsuperscript{\rm 2} Tencent Youtu Lab
    \textsuperscript{\rm 3} Xiamen University
    \\
    \{yutinggao, tristanli, sevjiang, scorpioguo, winfredsun\}@tencent.com\\ wangjp23@mail2.sysu.edu.cn,  hujguo@mail.sysu.edu.cn,  rrji@xmu.edu.cn
}

\begin{document}

\maketitle

\begin{abstract}
One significant factor we expect the video representation learning to capture, especially in contrast with the image representation learning, is the object motion. However, we found that in the current mainstream video datasets, some action categories are highly related with the scene where the action happens, making the model tend to degrade to a solution where only the scene information is encoded. For example, a trained model may predict a video as playing football simply because it sees the field, neglecting that the subject is dancing as a cheerleader on the field. This is against our original intention towards the video representation learning and may bring scene bias on a different dataset that can not be ignored. In order to tackle this problem, we propose to decouple the scene and the motion (DSM) with two simple operations, so that the model attention towards the motion information is better paid. Specifically, we construct a positive clip and a negative clip for each video. Compared to the original video, the positive/negative is motion-untouched/broken but scene-broken/untouched by \emph{Spatial Local Disturbance} and \emph{Temporal Local Disturbance}. Our objective is to pull the positive closer while pushing the negative farther to the original clip in the latent space. In this way, the impact of the scene is weakened while the temporal sensitivity of the network is further enhanced. We conduct experiments on two tasks with various backbones and different pre-training datasets, and find that our method surpass the SOTA methods with a remarkable 8.1\% and 8.8\% improvement towards action recognition task on the UCF101 and HMDB51 datasets respectively using the same backbone.

\end{abstract}

\section{Introduction}

\begin{figure}[t]
	\centering
	\includegraphics[width=.85\linewidth]{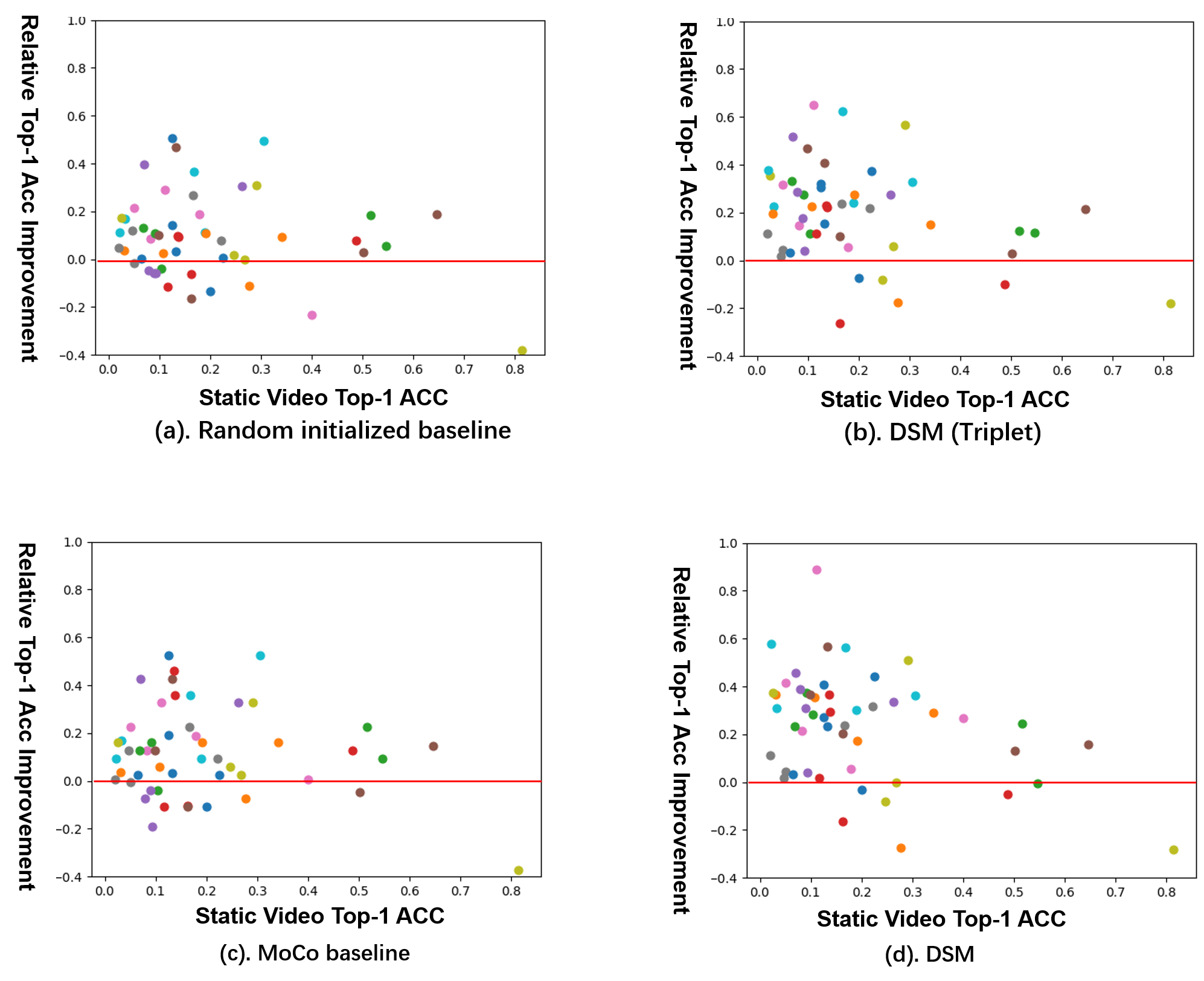}
	\caption{
Relative Top-1 accuracy improvement using the normal video over the artificially constructed static video. Each dot represents a semantic category. It can be seen from (a) that for some of the categories, \emph{i.e.}, dots below and around the red line, using a normal video does not help the model to give better prediction. This means that the scene and the motion are coupled together and models can give correct prediction by merely learning the scene. This phenomenon is relieved by our DSM method in (b) and (d), where the average relative improvement is notably lifted, meaning that the motion information is better exploited by the model. For a fair comparison, a MoCo baseline is given in (c).
    }
	\label{fig:teaser}
\end{figure}

Unsupervised representation learning has received widespread attention in the last few years. In the field of image representation learning, recent approaches\cite{he2020momentum, chen2020simple} have nearly surpassed their supervised counterparts. Nevertheless, in the field of video representation learning, there still exists a gap between unsupervised methods and supervised methods. In contrast with the image representation learning, one significant factor we expect video representation learning to capture is the \emph{object motion}, since a video usually contains continuous states of an object. While most commonly used image classification datasets are \textit{object-dominated}, \emph{i.e.}, object occupies a major part of the image, video datasets are usually \textit{scene-dominated}, \emph{i.e.}, object is relatively small and the discriminative information contained in the object motion is sometimes overwhelmed by statistics of the scene. To verify the side effect of this phenomenon, we conduct an experiment on the HMDB51\cite{kuehne2011hmdb} dataset by training models with two types of clips. One type of the clips are formed by repeating a frame that is randomly selected from a video, which have lost motion information and only the scene and static status are left. The other type of the clips are normal videos, which contains motion as expected. During testing, all samples are normal clips and results are shown in Figure \ref{fig:teaser}. Generally, we would expect all categories to show up upon the red line, meaning that motions are one of the key factors for video representation learning. However, we find that there are nearly 24\% of the categories show 
less than 5\% or no improvement with the help of motion information. This phenomenon may cause the model being lazy and only learn the scene without paying attention to the motion patterns that are what really matters. At the same time, certain actions in some datasets only occur in specific scenes, making the model prone to couple the motion pattern with the scene tightly. For example, a trained model may binding \textit{squats} to \textit{gyms}, thus misjudges when \textit{squats} happen in other scenes. It means that although scene and motion do promote each other sometimes, a strong couping between the two may make the learned representations generalize poorly and are easy to overfit to a specific training set.

To alleviate the scene bias problem, many efforts have been paid in the supervised setting. Simonyan et al.\citeyearpar{simonyan2014two} and Feichtenhofer et al.\citeyearpar{feichtenhofer2019slowfast} use a two-way convolutional neural network to capture appearance feature and temporal characteristic respectively at a cost of computation complexity. Zhao et al.\citeyearpar{zhao2018recognize} propose a new ConvNet architecture which can derive disentangled components, \emph{i.e.}, static appearance, apparent motion and appearance changes, from low-level visual feature maps. Girdhar et al.\citeyearpar{girdhar2020cater} build a synthetic video dataset with observable and controllable scene bias, forcing the model to understand both the spatial and temporal information to give correct prediction. 
Choi et al.\citeyearpar{choi2019can} propose to mitigate scene bias by augmenting the standard cross-entropy loss with an adversarial loss for scene type and a confusion loss of human mask.
Wang et al.\citeyearpar{Wang_2018_CVPR} explicitly pulls actions from context through an auxiliary two class classifier.

In this work, we try to alleviate the \textit{scene-dominated bias} in an unsupervised manner and propose to decouple the scene and the motion (DSM) with two simple operations. Specifically, we formulate self-supervised video representation learning as a data-driven metric learning  problem and construct a positive clip and a negative clip for each video. Compared to the original video, the positive/negative is motion-untouched/broken but scene-broken/untouched by \emph{Spatial Local Disturbance} and \emph{Temporal Local Disturbance}. Our objective is to pull the positive closer while pushing the negative farther to the original clip in the latent space. In this way, our model is more scene independent and more  motion sensitive. As shown in Figure \ref{fig:teaser}(b) and (d), our method notably improve the overall feature representation ability, especially for categories that strongly rely on temporal information.


Our contributions are summarized as follows:

\begin{itemize}[leftmargin=20pt,topsep=0pt]
\item We formulate the self-supervised video representation learning into a data-driven metric learning, and decouple the scene and the motion to alleviate the negative impact of the scene and the motion coupling problem which is commonly seen in the current video datasets.

\item We design two effective strategies to construct positive and negative sample pairs which consider both the spatial and the temporal characters of the video data.

\item Our method greatly improves the performance of unsupervised video representation learning and achieves state-of-the-art results on both UCF101\cite{soomro2012ucf101} and HMDB51\cite{kuehne2011hmdb} datasets. 

\end{itemize}

\section{Related Work}
\subsection{Video Representation Learning}

The most significant characteristic of the video representation learning is the requirement for temporal modeling compared to the image representation learning. Early works first use 2D CNNs to capture appearance features at the frame level and then do average pooling or adopt LSTM over the temporal dimension to learn motion patterns\cite{wang2018temporal,zhou2018temporal}. Another type of method\cite{simonyan2014two} use a two-way ConvNet to capture spatial appearance features and temporal motion patterns respectively. As a natural evolution, 3D CNNs \cite{tran2015learning, carreira2017quo,hara2018can} 
are later used to capture spatio-temporal patterns at the same time. Feichtenhofer et al.\citeyearpar{feichtenhofer2019slowfast} uses two-pathway on the basis of 3D network and achieves good results.
However, Li et al.\citeyearpar{Li_2018_ECCV} and Girdhar et al.\citeyearpar{girdhar2020cater} point out that the current commonly used video datasets are plagued with implicit biases over scene and object structure
and they propose two datasets, which requires a complete understanding of spatio-temporal information for a model to give correct prediction. Choi et al.\citeyearpar{choi2019can} and Wang et al.\citeyearpar{Wang_2018_CVPR} propose to mitigate scene bias from the perspective of training strategy.

\subsection{Self-supervised Learning}
Self-supervised learning has received extensive attention in the field of image classification. One common way is to define a pretext that are related to downstream tasks\cite{noroozi2016unsupervised,gidaris2018unsupervised,jenni2020steering}.
Another mainstream type is based on metric learning, which aims to minimize the distance between similar samples while pushing away dissimilar samples in the feature space.
Contrastive loss\cite{hadsell2006dimensionality} proposes to decrease the distance between positive pairs while pushing the negative pairs to a certain margin. Triplet loss\cite{schroff2015facenet} makes further improvements by introducing triplets, which minimizes the distance between an anchor and a positive sample and maximizes the distance between the anchor and a negative sample. Recent works based on contrastive loss \cite{wu2018unsupervised,he2020momentum,chen2020simple} have achieved excellent results on multiple visual tasks and narrowed the gap between the supervised learning and the unsupervised learning. 
The core idea in contrastive learning is to strengthen the invariance of the network to various data augmentations. In this article, we explore video representation learning under the framework of self-supervised learning.


\subsection{Self-supervised Video Representation Learning}

Recently, many self-supervised video representation learning methods have been proposed. Among them, one prominent direction is to design a surrogate signal that can been used as supervision such as sequence order of frames\cite{misra2016shuffle}, space-time cubic puzzles \cite{kim2019self}, video clip order \cite{xu2019self,luo2020video} and video playback rating \cite{yao2020video,benaim2020speednet}. 
Besides, Gan et al.\citeyearpar{gan2018geometry} and Wang et al.\citeyearpar{wang2019self2} use the statistics of optical flow as supervision.
Another mainstream category is based on contrastive learning, whose core is to construct positive samples and negative samples. TCN\cite{sermanet2018time} treats the same actions under different cameras as positive samples and different time periods of the same video as negative samples. IIC\cite{tao2020selfsupervised} regards multi-modal data as positive samples, and videos with shuffled frame order as negative samples. CVRL\cite{qian2020Spatiotempora} proposes temporally consistent spatial augmentation with simple operations and treats the generated results as positive samples. In this work, we design two simple but effective strategies to construct positive and negative samples.



\section{Methodology}

We address video representation learning in a self-supervised manner, and the core idea is to learn an embedding space in which temporally similar/dissimilar but context-variant/-invariant video clips are close/far. In particular, we propose two simple but powerful augmentation strategies to construct clip pairs from the spatio-temporal structure of a single video. In the following sections, we first give an overview of the entire architecture, and then introduce the augmentation methods and objective functions in details.

\begin{figure}
	\centering
	\includegraphics[width=.9\linewidth]{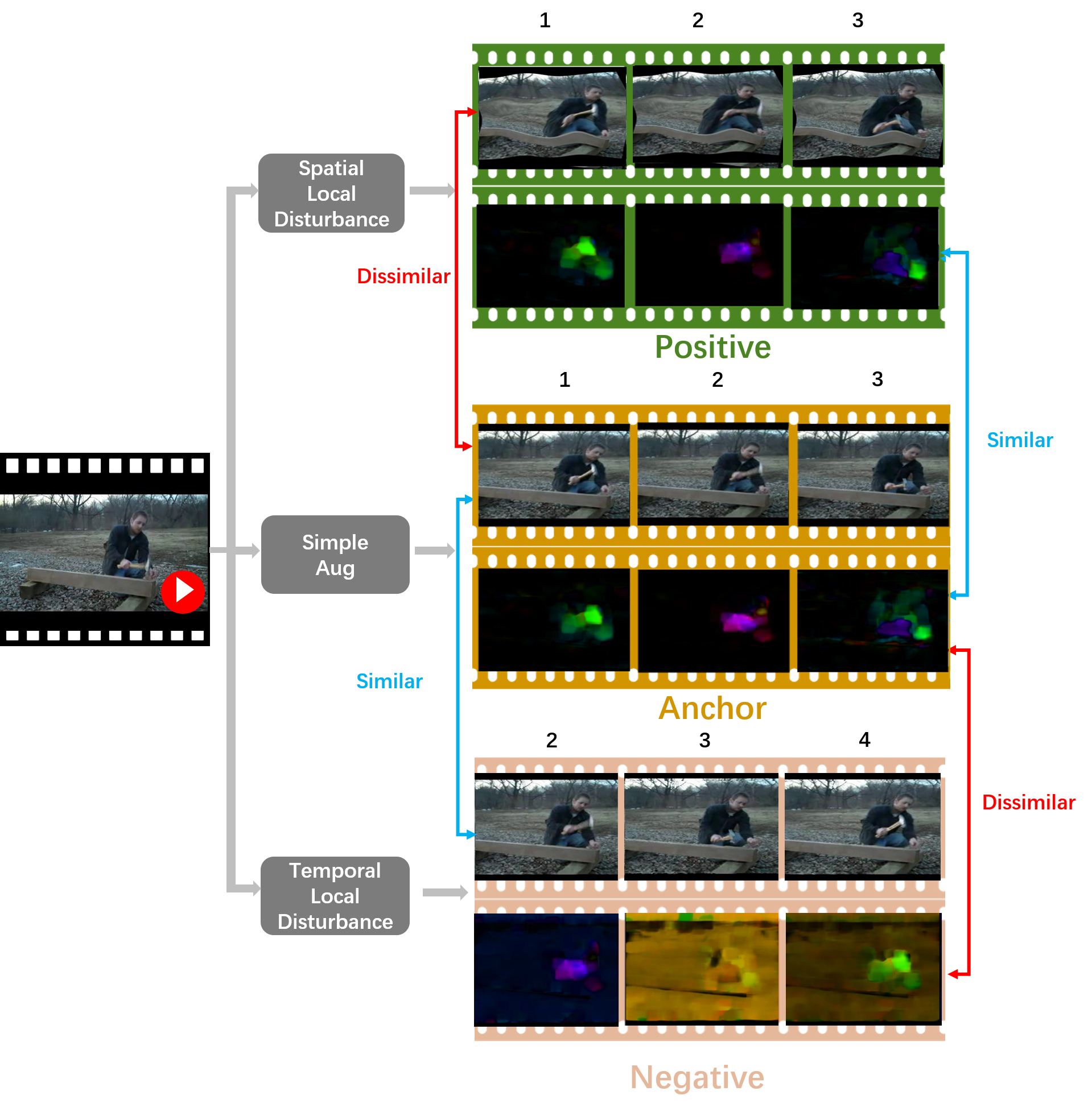}
	\caption{
	\textbf{Overview of our method}. We first construct positive and negative samples. Positive samples are constructed by \emph{Spatial Local Disturbance} while negtive samples are constructed by \emph{Temporal Local Disturbance}. Then we propose to learn video representations by training a convnet to push away temporal-dissimilar/spatial-similar pairs but pulling temporal-similar/spatial-dissimilar pairs closer. 
    }
	\label{fig:overall}
\end{figure}




\subsection{Overall Architecture}
\label{sec:overall}
The entire framework is shown in Figure \ref{fig:overall}. 
Formally, given an unlabeled video dataset $X$ that contains $N$ videos, we sample $T$ frames from the video for each clip and input the clip to the network.  
Random cropping is performed on each input to generate three clips with different spatial regions but maintaining temporal consistency, denoted as $c_{1}$, $c_{2}$ and $c_{3}$. 
Afterwards, we apply basic augmentation $b$, spatial warping $s$ and motion disturbance $t$ on these three clips respectively to construct a triplet, \emph{i.e.}, anchor ${a}=b(c_{1})$, positive sample ${p}=s(c_{2})$ and negative sample ${n}=t(c_{3})$. 
Compared to $a$, $p$ destroys the structural information of scene but maintains temporal semantics. Meanwhile, $n$ disturbs the local motion pattern of the moving subject 
but retains the spatial information. This triplet is fed into a 3D backbone $f$ to extract spatio-temporal features which are then projected to a D-dimension feature space followed by L2 normalization, and we denote the decoded feature as $z_a$, $z_p$ and $z_n$.
In this way, the triplet is projected into a normalized embedding space $(z_a, z_p, z_n) \in \mathcal{R}^D$. We then perform spatio-temporal representation learning in the normalized embedding space using two metrics: intra-video triple loss and contrastive loss, which will be introduced in details in the later section.

\subsection{Spatial Local Disturbance} \
\label{sec:positive_selection}
The core idea of positive sample construction is to break local contexts while keeping motion semantics basically unchanged with data augmentation. 
Spatial data augmentations, \emph{e.g.}, rotation, color jittering, have been widely used in the image-related task. However, it is underexplored in the video domain. A naive way for video augmentation may be applying existing image spatial augmentations to each frame of the video. However, some of these operation may damage the motion semantic. For example, if different rotation angles are used for consecutive frames, the generated video will looks like suffering a severe camera shake, making the video difficult to recognize. In order to make the temporal abstraction of the entire video remains similar, all consecutive frames of a video should perform the same transformation, and a video data augmentation that meets this requirement is \emph{temporally consistent}.

\begin{figure}
	\centering
	\includegraphics[width=\linewidth]{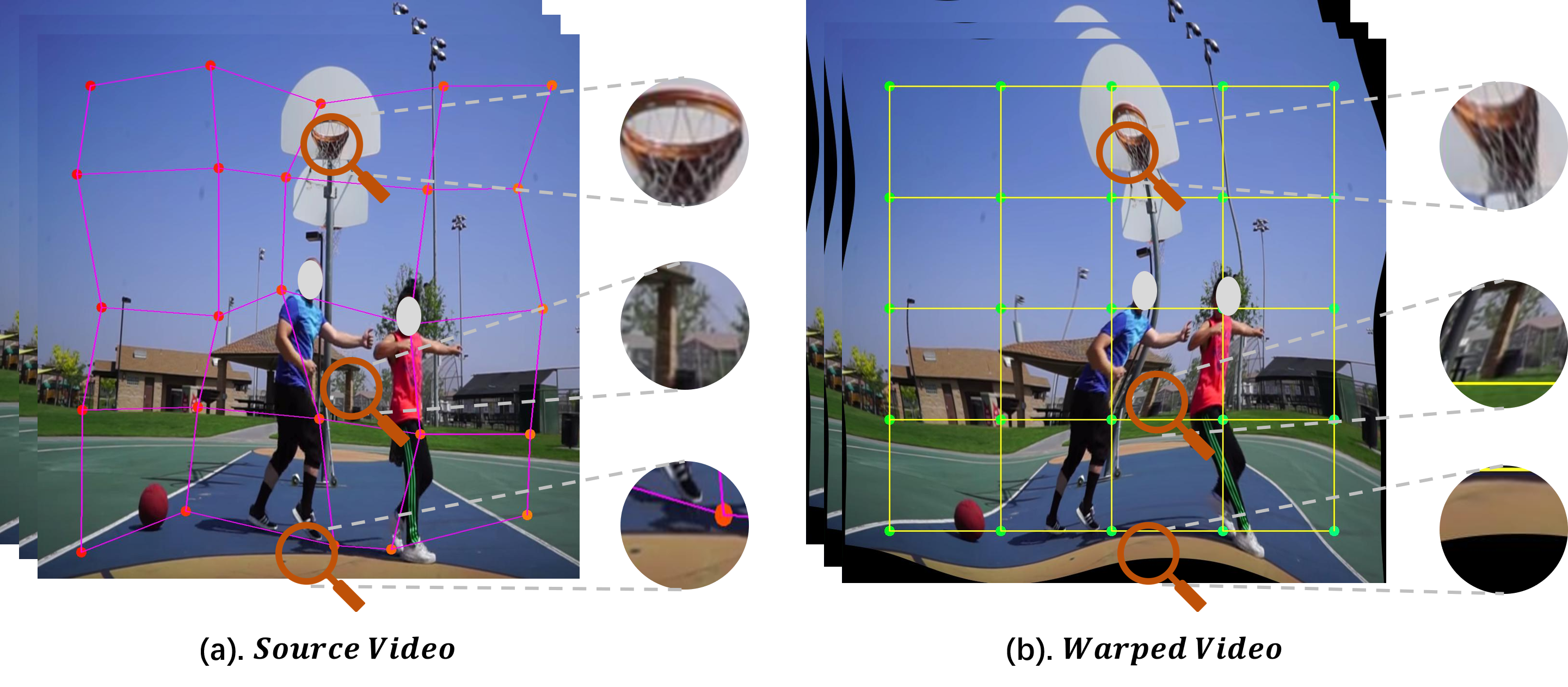}
	\caption{
	Illustration of the \textbf{Spatial Warping}, which randomly warps spatial regions in each epoch. Though the local statistics is broken, the global statistics is maintained.
    }
	\label{fig:spatial_warpping}
\end{figure}

Thin-Plate-Spline (TPS) is widely used in the OCR field to rectify distorted text regions\cite{jaderberg2015spatial,shi2016robust}.
In contrast, we aim to damage the statistics of the scene but keep motion pattern unchanged with TPS. Specifically, we select $N$ uniformly distributed destination points on the target video as $\mathcal{D}=\{d_i\}_{i=1,...,N} \in \mathbf{R}^{2 \times N}$.
For each destination point $d_i$, we add a small offset $\Delta_i=[\Delta x_{i}, \Delta y_{i}]^T$ to generate a corresponding source point $s_{i}=d_i + \Delta_i$ on the original video, making up a source point set $\mathcal{S}$. Both the horizontal and vertical offsets are randomly sampled from $[-C, C]$, where $C$ is typically one-tenth of the frame size. 
Then $s$ and $d$ are used to compute the parameters of TPS.
At last, the grid $P$ is computed by the TPS transformation and the final warped video is generated by a bilinear sampler given $P$ and the original video.
An illustration of spatial warping is shown in Figure \ref{fig:spatial_warpping}. In this way, although spatial local statistics are modified, the global context is maintained. 
In each training epoch, due to the randomness of the grid, the generated warped videos are different and the pixels in the local area always show huge differences from the original videos. Therefore, the network needs to focus more on motion pattern and pay less attention to spatial changes to extract consistent representations for the original and warped videos. Since all frames in the video perform the same transformation operation, it is a \textit{temporally consistent} augmentation.



\subsection{Temporal Local Disturbance}

In contrast to ${p}$, the major difference between ${a}$ and ${n}$ is the motion pattern.
A straightforward idea may be using other videos directly as ${n}$ as in recent contrastive learning method \cite{sermanet2018time}. However, besides the temporal information, there still exists many artificial cues to distinguish two videos, which are easier to solve for the network \cite{kim2019self}.
Therefore, it is not guaranteed that the network will focus on the motion.
To overcome this limitation, we propose to generate ${n}$ with large temporal abstraction differences but similar context from ${a}$ using \textit{Temporal Local Disturbance} (TLD). TLD comprises two transformations and are described in details as below. 

\noindent\textbf{Optical-flow Scaling.} 
We first denote a video as $I(x,y,t)$, where $x,y$ are spatial coordinates and $t$ is time. Under the brightness constancy assumption \cite{horn1981determining}, the relation between $I$ and the corresponding optical flow $(V_x,V_y)$ can be formulated as:
\begin{equation}
    \frac{\partial I}{\partial x}V_x + \frac{\partial I}{\partial y}V_y + \frac{\partial I}{\partial t} = 0,
\end{equation}
where $V_x$ and $V_y$ are the horizontal and vertical components of the velocity.
It should be noted that given a video frame at time $t$ and the corresponding optical flow, we can compute the $t+1$ frame by:
\begin{equation}
    I(x,y,t+1)=I(x+V_x,y+V_y,t)
\end{equation}

By applying the equation, we can accelerate or decelerate the video motion without changing background pixels too much. In particular, given a scale factor $\phi(t)$, we have
\begin{equation}
    \hat{I}(x,y,t+1) = I(x+ \phi(t) V_x, y + \phi (t) V_y,t) 
\end{equation}
where $\phi(t)$ is randomly sampled from [0, M] over time $t$, and $M$ is a hyperparameter controlling the amplitude. We find that a too big $M$ may result in unnatural videos with wide black boundary, and setting $M$ to 5 gives best result. 


\noindent\textbf{Temporal Shift.}
The purpose of Temporal Shift is to distinguish the temporal differences of various videos that contain similar scene. We assume that a video and its corresponding temporal shifted version has different motion patterns. Given a video $x$, we randomly and uniformly sample a shift scalar $\tau$ from $[ \alpha _1, \alpha _2]$, then the new video is generated by:
\begin{equation}
    \hat{x}_i = x_{i+\tau}, i \in 1,2...T
\end{equation}
That is, we extend the origin ${a}$ from indexs $1:T$ to ${n}$  with indexs $1+\tau:T+\tau$.
If the index of $\hat{x}_i$ exceeds the length of the untrimmed video, we loop the index from the beginning.

Intuitively, the choose of $\tau$ determines the similarity between $x_i$ and $\hat{x}_i$.
When $\tau$ approaches zero, the generated $\hat{x}$ looks very similar with $x$. 
To differentiate the ${n}$ and ${a}$ apart, the encoder network must focus on global rather local statistics.

\subsection{Objective Function}
We employ two objective functions to optimize the model. One is intra-video triplet learning, which generates negative sample by itself. The other one is the contrastive learning, which takes in other video as negative samples. While contrastive loss has been widely and successfully used in self-supervised methods, we verify that our methods does not rely on such loss design and generalize well with triplet loss.
\subsubsection{Intra-video Triplet Learning.}
We first optimize the network with the following objective function in the form of triplet loss.
\begin{equation}
    \label{eq:1}
    \mathcal{L}_t = \sum_{i=1}^{N} max \{d(z_{a_i}, z_{p_i}) - d(z_{a_i}, z_{n_i}) + margin, 0\}
\end{equation}
where $d(z_a, z_p) = ||z_a-z_p||_2$, $d(z_a, z_n) = ||z_a-z_n||_2$ and $margin$ is a hyperparameter to restrain the distance between $d(z_a, z_p)$ and $d(z_a, z_n)$.

\subsubsection{Contrastive Learning.}
Contrastive learning, \emph{e.g.}, InfoNCE \cite{hjelm2018learning}, learns to obtain representations by maximizing similarity of similar pairs over dissimilar pairs. Given a query $q$, a corresponding positive sample $k^+$ and other negatives $\{ k^-\}$, InfoNCE defines the contrastive loss as follows:
\begin{equation}
    \label{eq:2}
    \mathcal{L} = -log\frac{exp(q \cdot k_+  / \tau)}{exp(q \cdot k_+  / \tau) + \sum_{k^-}^{}exp(q \cdot k^- / \tau)}
\end{equation}
where $\tau$ is a temperature hyperparameter that is used to scale the distribution of distance. 
For simplicity, we set $\tau = 1$ by default. Then we introduce $(z_a, z_p, z_n)$ into InfoNCE and the final objective function is as follows:
\begin{equation}
    \label{eq:3}
    \begin{split}
    \mathcal{L}_c = -log\sum_{i=1}^{N}\frac{exp(z_{a_i} \cdot z_{p_i} )}{sim(z_{a_i},z_{p_i},z_{n_i})+ \sum_{j=0}^{K}exp(z_{a_i} \cdot z_{a_j})}
    \end{split}
\end{equation}
where $sim(z_{a_i},z_{p_i},z_{n_i}) = exp(z_{a_i}\cdot z_{p_i}) + exp(z_{a_i}\cdot z_{n_i})$ and $K$ is the number other samples. We use a memory bank with size $K$ to save features of ${a}$. Compared to the intra-video triplet, inter-video samples are also used as negative. 
We adopt MoCo \cite{he2020momentum} as the basic framework of contrastive representation learning for its efficacy and efficiency.

It can be seen from the equation \ref{eq:1} and equation \ref{eq:3} that the learning of embedding space depends on the quality of the generated positive and negative samples. By applying DSM, we expect the prediction not to be determined by the spatial context and take more motion pattern into account.
\section{Experiments}

\subsection{Implementation Details}

\textbf{Datasets.}
All the experiments are conducted on three video classification benchmarks, UCF101, HMDB51 and Kinetics \cite{kay2017kinetics}. 
UCF101 consists of 13,320 manually labeled videos in 101 action categories and HMDB51 comprises 6,766 manually labeled clips in 51 categories, both of which are divided into three train/test splits.
Kinetics is a large scale action recognition dataset that contains 246k/20k train/val video clips of 400 classes.


\noindent\textbf{Networks.}
We use C3D\cite{tran2015learning}, I3D\cite{carreira2017quo} and 3D ResNet-34\cite{hara2018can} as base encoders followed by a global average pooling layer and a fully connected layer to project the representations into a 128-dimensional latent space.

\noindent\textbf{Default Settings.}
All the experiments are conducted on 16 Tesla V100 GPUs with a batch size of 128. For each video clip, we uniformly sample 16 frames with a temporal stride of 4 and then resize the sampled clip to 16 $\times$ 3 $\times$ 224 $\times$ 224. The margin of triplet loss is set to 0.5 and the smoothing coefficient $m$ of momentum encoder in contrastive representation learning is set to 0.99 following MoCo\cite{he2020momentum}.
The memory bank size $K$ is set to 6536.
The boundary of temporal shift operation $\alpha_1$ is 2 and $\alpha_2$ is 20 for UCF101. Since the average length of Kinetics is larger than UCF101, $\alpha_1$ is set to 4 and $\alpha_2$ is set to 30.

\noindent\textbf{Pre-training Settings.}
We pre-trained the network for 200 epochs and adopt SGD as our optimizer with a momentum of 0.9 and a weight decay of 5e-4. The learning rate is initialized as 0.003 and decreases to 1/10 at the 80, 120 and 160 epoch.  

\noindent\textbf{Fine-tuning Settings.}
After pre-training, we transfered the weights of base encoder network to two downstream tasks, action recognition and video retrieval. We fine-tuned on each dataset for 45 epochs. The learning rate is initialized as 0.1 and multiplied by 0.1 every 10 epochs. 

\noindent\textbf{Evaluation Settings.}
For action recognition, follow common practice \cite{wang2019self}, the final result of a video is the average of the results of 10 clips that uniformly sampled from it during testing time.

\begin{table*}
    \centering
    {
    \begin{tabular}{p{5.5cm}llllll}
    \toprule
    Method&Year &Resolution&Pretrained&Architecture& UCF101 & HMDB51 \\
    \midrule
    \multicolumn{6}{l}{\textbf{Supervised}}\\
    Random init&- &$224\times224$& -& I3D & 47.9 & 29.6 \\
    ImageNet inflated&- &$224\times224$& ImageNet & I3D & 67.1 & 42.5 \\
    Kinetics supervised&- &$224\times224$& Kinetics & I3D & 96.8 & 74.5\\
    \midrule
    \multicolumn{6}{l}{\textbf{Self-supervised}}\\
    Puzzle 
    \cite{kim2019self} &AAAI'19& $112\times112$ & UCF101 & C3D& 65.0&31.3 \\
    VCP \cite{luo2020video} &AAAI'20& $112\times112$ & UCF101 & C3D & 68.5&32.5  \\
    PRP \cite{yao2020video} &CVPR'20& $112\times112$&UCF101&C3D&69.1&34.5\\
    MoCo \cite{he2020momentum} $\Diamond$ &CVPR'20&$112\times112$&UCF101 & C3D&60.5&27.2 \\
    \hline
    \textbf{DSM (Triplet)} &-&$112\times112$& UCF101&C3D &68.4 &38.2\\
    \textbf{DSM} &-&$112\times112$& UCF101&C3D &70.3 &40.5\\
    \hline
    Clip Order~\cite{xu2019self}  &CVPR'19&$112\times112$&Kinetics& R(2+1)D & 72.4 & 30.9 \\
    DPC~\cite{han2019video} &ICCVW'19&$224\times224$&Kinetics & 3D-ResNet34 & 75.7 & 35.7 \\
    AoT~\cite{wei2018learning}  &CVPR'18&$224\times224$&Kinetics& T-CAM & 79.4 & -   \\
    SpeedNet \cite{benaim2020speednet} &CVPR'20&$224\times224$&Kinetics& I3D &  66.7 & 43.7 \\
    MoCo \cite{he2020momentum} $\Diamond$ &CVPR'20&$224\times224$&Kinetics & I3D& 62.3 & 36.5 \\
    \hline
    \textbf{DSM (Triplet)} &-&$224\times224$& Kinetics&I3D & 70.7&48.5 \\
    \textbf{DSM (Triplet)} &-&$224\times224$& Kinetics&3D-ResNet34 &76.9 &50.3 \\
    \textbf{DSM} &-&$224\times224$& Kinetics&I3D &74.8&52.5 \\
    \textbf{DSM} &-&$224\times224$& Kinetics&3D-ResNet34 &\textbf{78.2} & \textbf{52.8} \\
    \bottomrule
    \end{tabular}
    }
    \caption{The top-1 accuracy (\%) of our method compared with previous approaches on the UCF101 and HMDB51 dataset. DMS(Triplet) is optimized with triplet loss. All the accuracy is averaged over three splits and $\Diamond$ means a custom implementation.}
    \label{tab:sota_action_recognition_cmp}
\end{table*}

\subsection{Action Recognition}

\subsubsection{Transfer Learning.}
We fine-tuned the whole model on UCF101 and HMDB51 with all labels, and the results are shown in Table \ref{tab:sota_action_recognition_cmp}. 
We also compare the results of the I3D network pretrained with all labels of ImageNet and Kinetics in a supervised manner. 
It can be seen from the experimental results that the encoder pre-trained with DSM can significantly surpass the random initialized counterpart across various network architectures and benchmarks. 
Compared to SpeedNet, one of the state-of-the-art methods, DSM brings 8.1\% and 8.8\% improvement on UCF101 and HMDB51 respectively.
We can also observe that our approach surpasses all the other self-supervised methods on both UCF101 and HMDB51 datasets using the same backbone.

\subsubsection{Test with Limited Labeled Data.}

\begin{figure}
	\centering
	\includegraphics[height=.45\linewidth]{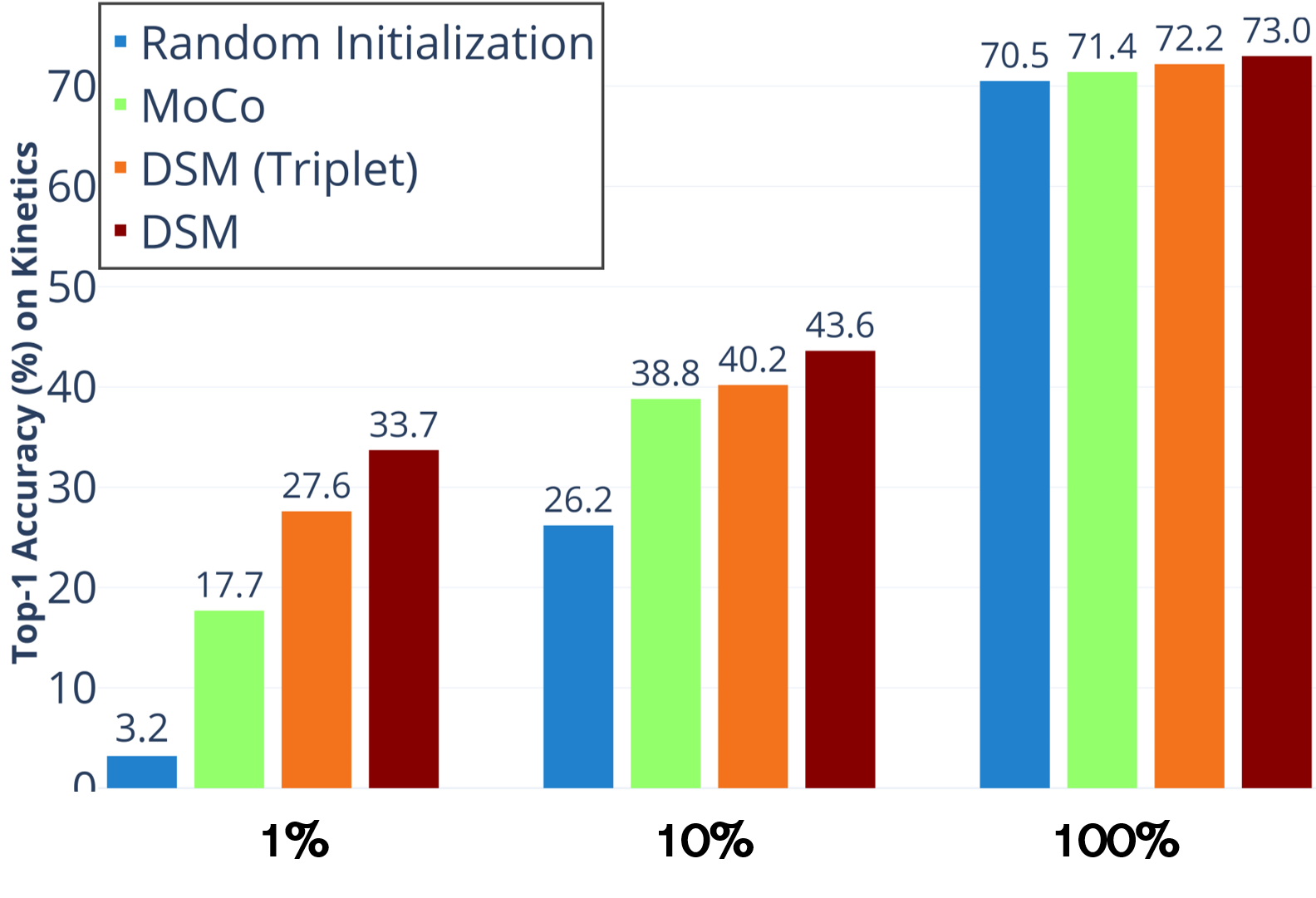}
	\caption{
	Top-1 accuracy with 1\%, 10\% and 100\% Kinetics labels on Kinetics. Our method brings prominent improvement especially with a small amount of labeled data. Compared to MoCo baseline, DSM brings an increase of 16\% with 1\% labels, an increase of 4.8\% with 10\% labels. When using a large amount of labeled data (100\% Kinetics labels), DSM can still increase the accuracy of MoCo baseline from 71.4\% to 73\%.}
	\label{fig:semi-supervised}
\end{figure}

Following SimCLR \cite{chen2020simple}, we sampled 1\% and 10\% labeled data from Kinetics in a class-balanced way ($\sim$6 and $\sim$57 videos per class) and fine-tuned the whole model with the sampled data. Figure \ref{fig:semi-supervised} exhibits the comparison results of our method with the MoCo baseline at 1\%, 10\% and 100\% labeled data on the validation set of Kinetics. We also report the results of the random initialized model for reference. All the experiments in this part use I3D as backbone. It can be seen from the figure that DSM significantly exceeds the MoCo baseline at all the volumes of labeled data. At the same time, DSM brings more prominent improvement with a small amount of labeled data. Specifically, with 1\% labeled data, the accuracy increases from 17.7\% to 33.7\%, and with 10\% labeled data, the accuracy increases from 38.8\% to 43.6\%. It is worth noting that the results of DSM(Triplet) also excel the MoCo baseline at each volume of the labeled data, and are quite close to DSM. This further proves that the intra-video positive and negative sample construction strategy is indeed effective. 
Moreover, when the amount of labeled data is very large, that is, using 100\% Kinetics labeled data, DSM can still increase the accuracy of MoCo baseline from 71.4\% to 73\%, which indicates that our method is well generalized.

\subsection{Video Retrieval}

\begin{table}
\footnotesize
\centering
\begin{tabular}{p{2.5cm}p{0.8cm}p{0.4cm}p{0.4cm}p{0.4cm}p{0.4cm}p{0.4cm}}
\toprule
\bf Method & \bf Net & \bf 1 & \bf 5 & \bf 10  & \bf 20 & \bf 50\\
\midrule
Jigsaw~\cite{jigsaw} & CFN  &  19.7 & 28.5 & 33.5 & 40.0 & 49.4 \\
OPN~\cite{opn} & OPN  & \textbf{19.9} & 28.7 &  34.0 &  40.6 &  51.6 \\
Clip Order~\cite{xu2019self} & C3D & 12.5 & 29.0 & 39.0 & 50.6 &  66.9 \\
Clip Order~\cite{xu2019self} & R3D & 14.1 & 30.3 & 40.0 & 51.1 & 66.5 \\
SpeedNet\cite{benaim2020speednet} & S3D-G & 13.0 & 28.1 & 37.5 &  49.5 & 65.0 \\
\midrule
\textbf{DSM} & C3D & 16.8 &33.4 &43.4 &54.6 & 70.7 \\
\textbf{DSM} & I3D & 17.4 & \textbf{35.2} & \textbf{45.3} & \textbf{57.8} & \textbf{74.0} \\
\bottomrule
\end{tabular}
\caption{{\bf Recall-at-topK (\%).} Accuracy under different K values on UCF101.}
\label{tab:recallatk_ucf101}
\end{table}

\begin{table}
\footnotesize
\centering
\begin{tabular}{p{2.5cm}p{0.8cm}p{0.4cm}p{0.4cm}p{0.4cm}p{0.4cm}p{0.4cm}}
\toprule
\bf Method & \bf Net & \bf 1 & \bf 5 & \bf 10  & \bf 20 & \bf 50\\
\midrule
Clip Order~\cite{xu2019self} & C3D & 7.4 & 22.6 & 34.4 & 48.5& 70.1 \\
Clip Order~\cite{xu2019self} & R3D & 7.6 & 22.9 & 34.4 &48.8 & 68.9 \\
VCP \cite{luo2020video} & C3D& 7.8 & 23.8 & 35.3 & 49.3 &71.6 \\
\midrule
\textbf{DSM} & C3D &\textbf{8.2} &\textbf{25.9} &\textbf{38.1} & 52.0 & 75.0 \\
\textbf{DSM} & I3D & 7.6 & 23.3 & 36.5 & \textbf{52.5} & \textbf{76.0} \\
\bottomrule
\end{tabular}
\caption{{\bf Recall-at-topK (\%).} Accuracy under different K values on HMDB51.}
\label{tab:recallatk_hmdb51}
\end{table}

In this section, we evaluated DSM on video retrieval task. Following Clip Order and SpeedNet, the network is fixed as a feature extractor after pre-training with DSM on the split 1 of UCF101. Then the videos from both UCF101 and HMDB51 are divided into clips in units of 16 frames.
All the clips in the training set constitute a \textit{Gallery}, and each clip in the test set is used as a \textit{query} to retrieve the most similar clip in the \textit{Gallery} with cosine distance. 
If the category of the query appears in the K-nearest neighbors is retrieved, then it is considered as a hit. 
It should be noted that in order to keep the scale of representations generated by each 3D architecture consistent, we replaced the original global average pooling with an adaptive max pooling, yielding representations with a fixed scale of $1024 \times 2 \times 7 \times 7$. We show the accuracy when $K = 1, 5, 10, 20, 50$ and compare with other self-supervised methods on UCF101 and HMDB51 in Table \ref{tab:recallatk_ucf101} and Table \ref{tab:recallatk_hmdb51} respectively. It can be seen that when using the same backbone C3D, DSM surpasses the mainstream method Clip Order on the UCF101, and surpasses Clip Order and VCP on the HMDB51, which proves that the representations extracted by DSM are more discriminative.




\subsection{Ablation Study}

In this section, we explore the effectiveness of each component in the DSM. 
Results are shown in Table \ref{tab:ablation_cmp}, from which we can conclude that all of these components lead to better results, and in the way of generating negative samples, both scaling optical-flow and temporal-shift are effective and the combination of the two can bring further gains. 

\begin{table}[t]
 \setlength{\belowcaptionskip}{1pt}
    \centering
    {
    \begin{tabular}{llll}
    \toprule
    Positive & Negative & UCF101 & HMDB51 \\
    \midrule
    \multicolumn{4}{l}{\textbf{MoCo baseline}}\\
    -& -& 62.3 & 36.5 \\
    \midrule
    \multicolumn{4}{l}{\textbf{Our Method}}\\
    S-warping & -& 66.4 \textcolor{GREEN}{(+4.1)} &41.3 \textcolor{GREEN}{(+4.8)} \\
    -& M-disturb & 67.7 \textcolor{GREEN}{(+5.4)} & 44.0 \textcolor{GREEN}{(+7.5)}\\
    S-warping & Scaling-Of & 69.4 \textcolor{GREEN}{(+7.1)} & 47.5 \textcolor{GREEN}{(+11.0)} \\
    S-warping & T-Shift & 71.2 \textcolor{GREEN}{(+8.9)} &50.4 \textcolor{GREEN}{(+13.9)} \\
    S-warping & M-disturb &74.8 \textcolor{GREEN}{(+12.5)} & 52.5 \textcolor{GREEN}{(+16.0)}\\
    \bottomrule
    \end{tabular}
    }
    \caption{Ablation study of each component on the UCF101 and HMDB51. 
    All the methods are pre-trained on Kinetics with I3D backbone.
    S-warpping, M-disturb, Scaling-Of and T-Shift denote spatial warpping, motional disturbance, scaling optical-flow and temporal-shift respectively.}
    \label{tab:ablation_cmp}
\end{table}


\subsection{Analysis}

\begin{figure}
	\centering
	\includegraphics[width=\linewidth]{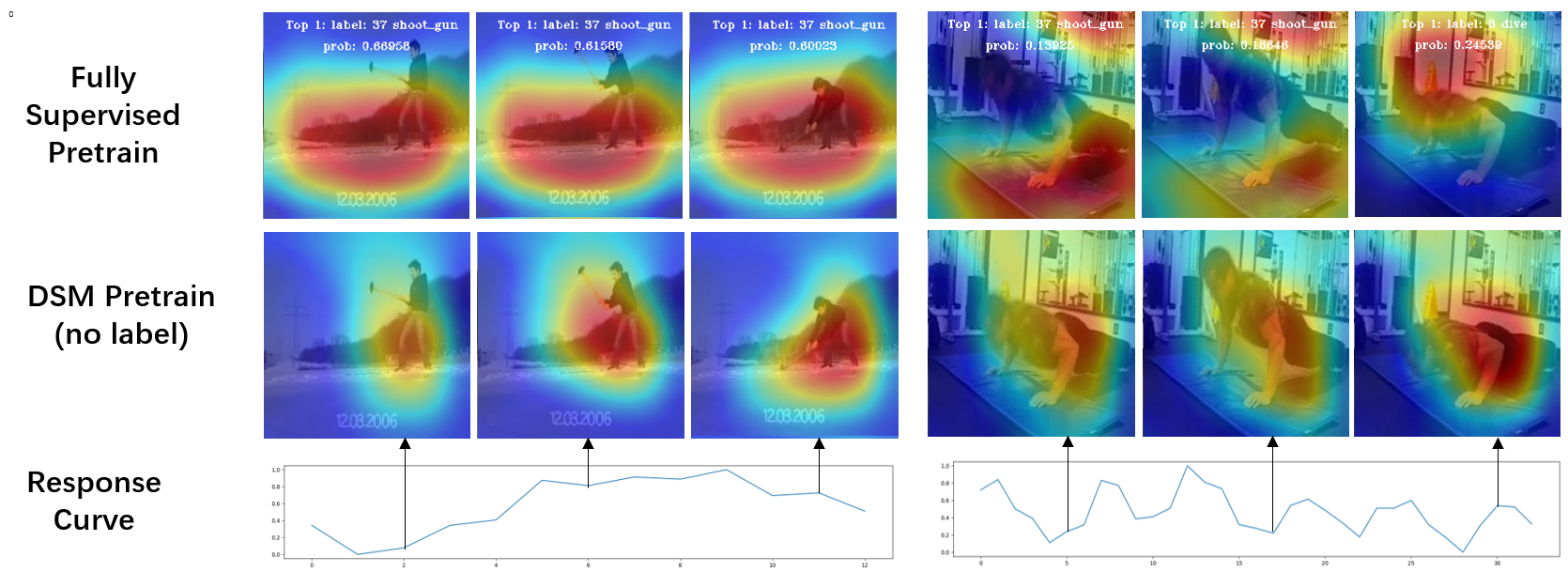}
	\caption{
	\textbf{Which space-time regions does the trained model focus on?}
	Notice that these action categories did not appear during the training time. The model trained with labels focus more on background and shows poor generation on new classes while the model pre-trained using DSM without any label pay more attention to the moving area. In addition, our method has a high response value for clips with strong motion information, and vice versa. 
	}
	\label{fig:heatmap}
\end{figure}

\textbf{Visualizing salient regions.}
In order to analyse which space-time regions our model focus on, we visualize the energy of the extracted representations with CAM\cite{zhou2016learning}.
For comparison, we pre-train two models using I3D as backbone on the UCF101 under two settings:
\textit{i.} fully supervised,
\textit{ii.} self-supervised using DSM. We select some videos with obvious motion from HMDB51 and use a sliding window to generate multiple clips for each video, and then visualize the corresponding activation maps of these clips in Figure \ref{fig:heatmap}. Specifically, assuming that each video has $L$ frames, a sliding window with a scale of 16 and a stride size of 4 slides on the temporal dimension, generating $(L-16)/4$ clips for each video. All clips of each video are input to the above two models and the extracted feature representations of the last 3D layer before global average pooling is of the shape of $1024 \times T \times N \times N$, where $T$ and $N$ are the scale of the temporal and spatial dimension. 
Afterwards, we average over all channels to compress these features into the shape of $T \times N \times N$, then upsample and mask these heatmaps to the original videos. Visualization results under setting \textit{i} are displayed in the first row, and those under setting \textit{ii} are shown in the second row. It can be observed that the supervised approach is severely affected by the scene bias and falsely focus on the static background. On the contrary, DSM suffer less from scene bias and correctly focus on moving objects. Moreover, for setting \textit{ii}, we average and normalize the feature of each clip into a scalar, which is recorded as response value, then we plot the curve of all clip response values over time. We find the curve has a low value when there is a inconspicuous movement, such as a clip that is about to end an action, and the alternate phase of a cyclic action, which is consistent with our original intention to enhance the temporal sensitivity of the model. 


\begin{figure}[t]
	\centering
	\includegraphics[width=\linewidth]{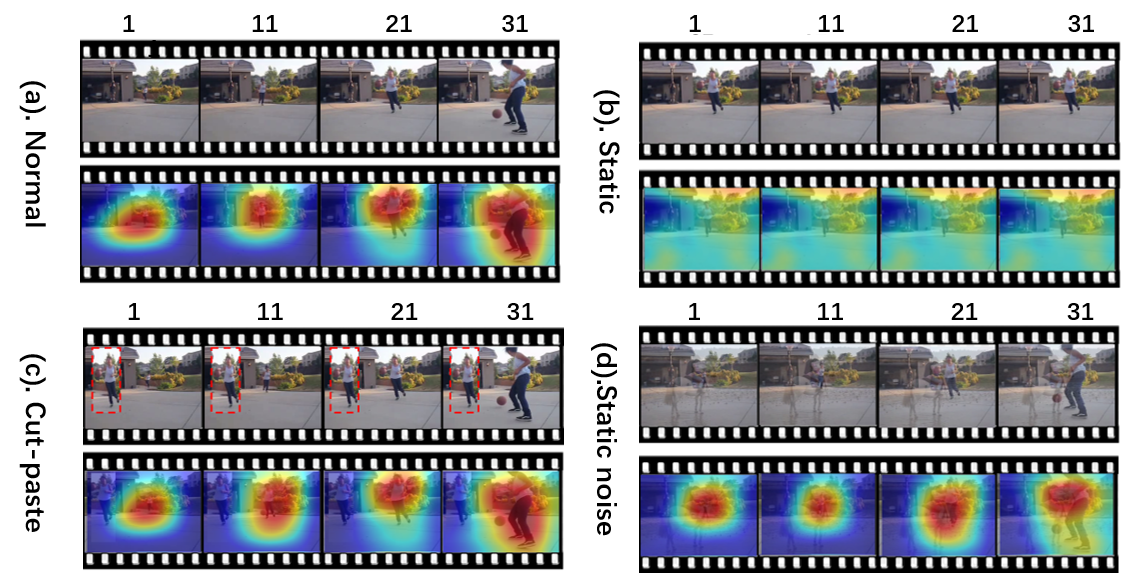}
	\caption{
	\textbf{Does the model really learn to focus on motion regions?} The video \textit{dribble} is selected from HMDB51 and the model is pre-trained with DSM on Kinetics. It can be concluded that: \emph{i}. When the input is a static video, DSM doesn't know where to focus on.
	\emph{ii}. When pasting another static human body, DSM still focus on the real movement area, which indicates that our method is even robust to human body distraction.
	\emph{iii}. Using a static frame as noise has no effect on our model.
    }
	\label{fig:where_to_look}
\end{figure}

\noindent\textbf{Adversarial examples.}
Since each action is carried out by a subject, a natural question comes out: does the model only learn to focus on the human body or it really learns to focus on the movement areas? To verify this question, we generate some adversarial samples in Figure \ref{fig:where_to_look}.
First, using a static video (copy one single frame multiple times) as input, the model shows random response. Then we paste another human body or introduce a static frame as noise, our method still correctly focus on movement area. The experiments in this part prove that the feature representation extracted by DSM have a fully understanding of space-time.

\section{Conclusion}

Due to the ubiquitously existing scene and motion coupling problem in the current video dataset, there are many actions that can be recognized simply from a static background. However, only focusing on the background does not generalize the model well in the open scene and may dwarft the temporal modeling. This paper presents DSM, a novel self-supervised method to overcome the influence of implicit bias over scenes. Combined with the metric learning, our method has a high tolerance towards the scene variants. We evaluate DSM both quantitatively and qualitatively. On the two popular benchmarks UCF101 and HMDB51, the proposed methods improves the state-of-the-art notably. And by visualizing the model focus map, our method correct focus on the motion instead of the unrelated area. We extend our method to retrieval and detection and good results are also achieved.

\bibliography{main.bib}

\end{document}